\documentclass[times,twocolumn,final]{elsarticle}

\usepackage{framed,multirow}

\input{preambule.sty}

\definecolor{newcolor}{rgb}{.8,.349,.1}

\journal{Medical Image Analysis}

\begin{document}


\begin{frontmatter}

\title{Model-Based Multi-Parameter Mapping}

\author[1,2]{Ya\"el Balbastre\corref{cor1}}
\cortext[cor1]{Corresponding author}
\author[1,3]{Mikael Brudfors}
\author[4]{Michela Azzarito}
\author[1]{Christian Lambert}
\author[1]{Martina F. Callaghan}
\author[1]{John Ashburner}

\address[1]{Wellcome Center for Human Neuroimaging, Queen Square Institute of Neurology, University College London, London, UK}
\address[2]{Athinoula A. Martinos Center for Biomedical Imaging, Massachusetts General Hospital and Harvard Medical School, Boston, USA}
\address[3]{School of Biomedical Engineering and Imaging Sciences, King's College London, London, UK}
\address[4]{Spinal Cord Injury Center Balgrist, University Hospital Zurich, University of Zurich, Switzerland}


\begin{abstract}
Quantitative MR imaging is increasingly favoured for its richer information content and standardised measures. However, computing quantitative parameter maps, such as those encoding longitudinal relaxation rate ($R_1$), apparent transverse relaxation rate ($R_2^\ast$) or magnetisation-transfer saturation (MT\textsubscript{sat}), involves inverting a highly non-linear function. Many methods for deriving parameter maps assume perfect measurements and do not consider how noise is propagated through the estimation procedure, resulting in needlessly noisy maps. Instead, we propose a probabilistic generative (forward) model of the entire dataset, which is formulated and inverted to jointly recover (log) parameter maps with a well-defined probabilistic interpretation (e.g., maximum likelihood or maximum a posteriori). The second order optimisation we propose for model fitting achieves rapid and stable convergence thanks to a novel approximate Hessian. We demonstrate the utility of our flexible framework in the context of recovering more accurate maps from data acquired using the popular multi-parameter mapping protocol. We also show how to incorporate a joint total variation prior to further decrease the noise in the maps, noting that the probabilistic formulation allows the uncertainty on the recovered parameter maps to be estimated. Our implementation uses a PyTorch backend and benefits from GPU acceleration. It is available at \url{https://github.com/balbasty/nitorch}.
\end{abstract}


\end{frontmatter}


\section{Introduction}
\label{sec:intro}

The magnetic resonance imaging (MRI) signal is governed by a number of tissue-specific parameters. While many common MR sequences only aim to maximise the contrast between tissues of interest, the field of quantitative MRI (qMRI) is concerned with the extraction of the original parameters \citep{Tofts2003}. This interest stems from the fundamental relationship that exists between the magnetic parameters and the tissue microstructure: the longitudinal relaxation rate $R_1=1/T_1$ is sensitive to myelin content \citep{Sigalovsky2006,Dick2012,Sereno2013,Lutti2014}; the apparent transverse relaxation rate $R_2^\ast=1/T_2^\ast$ can be used to probe iron content \citep{Ordidge1994,Ogg1999,Hasan2012,Langkammer2010}; the magnetization-transfer saturation ratio (MT\textsubscript{sat}) is related to protons bound to macromolecules (in contrast to free water) and offers another metric to investigate myelination \citep{Tofts2003b,Helms2008b,Campbell2018}. Additionally, quantitative MRI allows many of the scanner- and centre-specific effects to be factored out, making measures more comparable across sites \citep{Tofts2006,Deoni2008,Bauer2010,Weiskopf2013a}. 

\subsubsection*{Multi-parameter mapping}

Estimating quantitative parameters involves acquiring a collection of MR images while varying the sequence parameters (repetition time, echo time, flip angle) that are known to interact with the quantitative parameters, so that the function that governs the weighted signal intensity can be inverted. This function emerges from the integration of ordinary differential equations \citep{Bloch1946} and is therefore nonlinear and can be difficult to invert in closed-form. Therefore, early attempts at quantitative MRI targeted a single parameter at a time and used sequences with parameters carefully chosen so that other terms would disappear with a few algebraic manipulations \citep{Gupta1977}. However, as quantitative mapping moved from pure MR research to clinical applications, it became important to design sequences that allowed a maximum number of parameters to be estimated from a minimum number of acquisitions.

The multi-parameter mapping (MPM) protocol stems from this need: it uses multi-echo spoiled gradient echo (SPGR) acquisitions with variable flip angles and allows the quantification of $R_1$, $R_2^\ast$, MT\textsubscript{sat} and proton density (PD) at submillimetric resolution and in a clinically acceptable scan time \citep{Weiskopf2013a}. In this context, a wider range of sequence parameters can be used, at the price of rational approximations of the signal \citep{Helms2008b}. Images acquired with a SPGR sequence are parameterised by three parameters (PD, $R_1$, $R_2^\ast$) plus an eventual fourth one if MT weighting is used. The ESTATICS forward model reparameterises the SPGR signal equation using one \emph{intercept} per contrast (\emph{i.e.}, weighted volumes extrapolated to the theoretical \emph{zero} echo time) and a shared $R_2^\ast$ decay \citep{Weiskopf2014a, Mohammadi2017}. While ESTATICS provides a model-based estimation of $R_2^\ast$, the intercepts remain weighted images and another model must be inverted to recover the quantitative parameter maps: $R_1$, PD and MT\textsubscript{sat}\footnote{This procedure neglects noise propagation and assumes that successive quantities are exact -- even though they are not.}. While this inversion has an analytical solution when all contrasts have the same repetition time \citep{Wang1987, Liberman2014, Mohammadi2017}, rational approximations must be used as soon as different repetition times are used \citep{Helms2008c}. 

\subsubsection*{Non-linear fitting in MR relaxometry}

Many quantitative mapping methods discard the probabilistic nature of the acquired signal, which randomly deviates from its theoretical value due to thermal and physiological noise. This noise propagates through every computational step and ends up in the maps with a non-trivial distribution \citep{Polders2012, Cercignani2006b} and can even induce bias in the estimators \citep{Sijbers1999, Chang2008}. An alternative is to estimate well defined probabilistic quantities \emph{conditioned} on the observed data, such as posterior distributions or maximum-likelihood estimators \citep{Hurley2012, Tisdall2016} -- although they can still be biased \citep{Tisdall2017}. The use of qMRI estimators that directly optimise a probability distribution have remained elusive, in part due to the non-linear and non-convex nature of the corresponding log-likelihood. Still, several groups have proposed to fit quantitative maps using iterative methods, either from magnitude images \citep{Hurley2012, Tisdall2017} or directly from k-space measurements \citep{Chen1998, Sumpf2011, Sumpf2014, Zhao2014a}. In order to solve the non-linear system, \cite{Tisdall2017} used Brent's method, which does not require derivatives; \citet{Chen1998} used a Levenberg-Marquardt algorithm, \citet{Sumpf2011, Sumpf2014} used a conjugate gradient algorithm, while \citet{Zhao2014a} used a quasi-Newton Broyden-Flecher-Goldfarb-Shannon (BFGS) optimiser, which is not ensured to converge in general \citep{Mascarenhas2014} and is highly sensitive to initialisation. \citet{Scholand2020} pushed this idea further by modelling the entire Bloch equations and spatial encoding. In their work, the problem is solved using Gauss-Newton optimisation (which corresponds to Fisher's scoring of the Hessian). However, for this much more complex inverse-problem to be well-posed, spiral trajectories are needed so that each shot acquires as many spatial frequencies as possible. In multi-exponential $T_2$ mapping, the most common procedure involves non-negative least-squares fits \citep{Whittall1989}, which can be very sensitive to noise and optimisation parameters \citep{Wiggermann2020}. The use of non-linear least-squares algorithms is also common and revolves around (constrained) Gauss-Newton or Levenberg-Marquardt \citep{Milford2015, Gong2020a}. 


\subsubsection*{Denoising}

The push to higher resolutions and the use of parallel imaging reduces the signal-to-noise ratio (SNR) of the acquired images and consequently the precision of the computed parameter maps. \citet{Papp2016a} found a scan-rescan root mean squared error of about 7.5\% for $R_1$ at 1mm, in the absence of inter-scan movement. Smoothing can be used to improve SNR, but at the cost of lower spatial specificity. In this context, providing mapping methods that are robust to high levels of noise, both in terms of bias and variance, is of major importance.

Denoising methods aim to separate signal from noise. They take advantage of the fact that signal and noise have intrinsically different spatial profiles: thermal noise is spatially independent and often has a characteristic distribution while the signal is highly structured. Denoising methods originate from partial differential equations, adaptive filtering, variational optimisation or Markov random fields, and many connections exist between them. Two main families emerge:
\begin{enumerate}
    \item Inversion of a generative model\footnote{\emph{Generative model} should be understood in the classical sense: a (probabilistic) function that mimics the true data-generating process.}:
    $$\textstyle\hat{Y} = \argmin_Y \mathcal{E}_l\left(X, \mathcal{F}(Y)\right) + \mathcal{E}_p\left(\mathcal{G}(Y)\right),$$
    where $X$ is the observed data, $Y$ is the unknown noise-free data, $\mathcal{F}$ is an arbitrary \emph{forward} transformation (\emph{e.g.}, spatial transformation, downsampling, smoothing) mapping from the reconstructed to the observed data and $\mathcal{G}$ is a linear transformation (\emph{e.g.}, spatial gradients, Fourier transform, wavelet transform) that extracts features of interest from the reconstruction. $\mathcal{E}_l$ and $\mathcal{E}_p$ are two log-probability distributions that correspond to the likelihood of the parameters and their prior.
    \item Application of an adaptive nonlocal filter: 
    $$\textstyle\hat{Y}_i = \sum_{j\in\mathcal{N}_i} w\left(\mathcal{P}_i(X),\mathcal{P}_j(X)\right) X_j,$$
    where the reconstruction of a given voxel $i$ is a weighted average over all observed voxels $j$ in a given (possibly infinite) neighbourhood $\mathcal{N}_i$, with weights reflecting similarity between patches $\mathcal{P}$ centred about these voxels. Convolutional neural networks naturally extend this family by stacking many such filters and making their weights learnable rather than fixed.
\end{enumerate}
For the first family of methods, it was found that the denoising effect is stronger when $\mathcal{E}_p$ is an absolute norm (or sum of), rather than a squared norm, because the solution is implicitly sparse in the feature domain \citep{Bach2011}. This family of methods includes total variation (TV) regularisation \citep{Rudin1992a} and wavelet soft-thresholding \citep{Donoho1995}. The second family also leverages sparsity in the form of redundancy in the spatial domain; that is, the dictionary of patches necessary to reconstruct the noise-free images is smaller than the actual number of patches in the image. Several such methods have been developed specifically for MRI, with the aim of finding an optimal, voxel-wise weighting based on the noise distribution \citep{Coupe2008c,Manjon2010d,Coupe2012,Manjon2012b}.

Optimisation methods can naturally be interpreted as a maximum \emph{a posteriori} (MAP) solution in a generative model, which eases their interpretation and extension. This feature is especially important in qMRI, where we possess a well-defined (nonlinear) forward function and log-likelihood, and wish to regularise a small number of maps. Joint total variation (JTV) \citep{Sapiro1996} is an extension of TV to multi-contrast images, where the absolute norm is defined across contrasts, introducing an implicit correlation between them. TV and JTV have been used before in MR reconstruction, \emph{e.g.}, in compressed-sensing \citep{Huang2012}, quantitative susceptibility mapping \citep{Liu2011} and super-resolution \citep{Brudfors2018}. Recently, \citet{Varadarajan2020} applied JTV to multi-echo SPGR data for $B_0$ distortion correction, with the joint prior defined across echoes. However, these problems are linear and (therefore) convex, while the MR signal equations are not. This non-convexity has limited the use of iterative methods in this context, as algorithms either have a slow convergence -- in the case of first order methods -- or are not ensured to converge -- in the case of second order methods. 

\subsubsection*{Non-linear MPM fitting}

In \citet{Balbastre2020}, we introduced a non-linear, spatially regularised version of ESTATICS that was fit using Gauss-Newton optimisation. However, since regularisation did not apply directly to the parameter maps but to these weighted intercepts, the denoising of the quantitative parameters was merely a by-product of the denoising of the intercepts. In this work, we instead propose a generic framework that directly finds MAP quantitative (log) parameters from all SPGR data at hand. To this end, we extend our optimisation framework to work directly with the SPGR signal equation, and jointly regularise the log-parameter maps. Working with log-parameters $\log \text{PD}$, $\log R_1$, $\log R_2^\ast$, $\operatorname{logit} \text{MT}_{\text{sat}}$) is advantageous as their spatial gradients are independent from the choice of unit and representation (time, in $s$ or $ms$, or rate, in $s^{-1}$ or $ms^{-1}$), although it introduces an additional degree of nonlinearity. Another advantage is that posterior distributions are often better approximated by Laplace's method when a log basis is used \citep{MacKay1998b}. We name this extended model SPGR+JTV when it uses regularisation and SPGR-ML when it does not. In Gauss-Newton, positive-definiteness of the Hessian is enforced by the use of Fisher's scoring, which discards terms that involve second derivatives of the SPGR signal function. In this new work, we found that a more stable algorithm can be derived by keeping the (absolute) diagonal of this second-order term. We illustrate the improved convergence on a toy problem, and experimentally show that it generalises to least-squares ESTATICS and SPGR. Additionally, our implementation uses a quadratic upper bound of the JTV functional, allowing the surrogate problem to be efficiently solved using second-order optimisation.

We validated our method on a unique dataset -- five repeats of the MPM protocol acquired, within a single session, on a healthy subject -- by leave-one-out cross-validation: maps were estimated on one repeat and used to predict the weighted images from the other repeats. Our methods (SPGR+JTV and SPGR-ML) were compared to algorithms that rely on different flavours of the ESTATICS model: LOGLIN uses a log-linear fit to recover the intercepts and decay \citep{Weiskopf2014a}, whereas NONLIN, and NONLIN+JTV use a (regularised) non-linear fit \citep{Balbastre2020}. We also applied ESTATICS to echoes that had been previously denoised using the adaptive optimized nonlocal means (AONLM) method \citep{Manjon2010d}, as a baseline for the denoising component of our method. Quantitative maps were then computed analytically from the intercepts obtained from all ESTATICS methods. SPGR+JTV consistently outperformed all baselines. Besides, we probed the effect of regularisation by reconstructing maps with a varying number of echoes, and explored the use of our generative model to estimate uncertainty about the inferred quantitative values.


\section{Theory}
\label{sec:theory}

\subsection{Forward model: Spoiled Gradient Echo signal equation}
\label{sub:signal}

The steady state signal intensity of a spoiled gradient echo acquired with flip angle $\alpha$, repetition time TR and echo time TE is\footnote{A spoiled gradient echo acquisition is obtained by playing an excitation pulse that causes the bulk magnetization to rotate by an angle $\alpha$ (the flip angle), followed by a series of dephasing and rephasing gradients that create an echo train.}
\begin{equation}
    S = A \sin\alpha \frac{1 - \exp\left(-R_1 \text{TR}\right)}{1 - \cos\alpha\exp\left(-R_1 \text{TR}\right)}
    \exp\left(-R_2^\ast \text{TE}\right) ~,
\end{equation}
where $A$ is proportional to the proton density (with a multiplicative factor related to the sensitivity of the receiver coil).

Multi-echo techniques acquire multiple echoes, allowing the $R_2^\ast$ decay rate to be mapped, while variable flip-angle (VFA) techniques acquire images with multiple flip-angles to map the $R_1$ relaxation rate. Multi-parameter mapping protocols combine these two techniques in a single imaging session.

In addition, a similar sequence can be used to measure the magnetisation transfer saturation: in a two-pool model, protons are separated between the free pool -- protons in free water that can diffuse freely, which cause the MR signal -- and bound pool -- non-acqueous protons, \emph{e.g.}, in proteins and lipids of the myelin sheath encasing axons. Following selective saturation of the bound pool via an off-resonance pulse, the process of magnetisation transfer leads to partial saturation of the observable free pool. By acquiring images with and without this off-resonance pulse, the proportion of saturated protons can be measured, giving a surrogate measure of the bound pool. The resulting signal (which we write at TE=0 for brevity) can be modelled by adding an extra excitation pulse, with repetition time TR$_{2}$ and flip angle $\alpha_2$ \citep{Helms2008b}:\\
\begin{equation}
    S_{\text{TE}=0} = A \sin\alpha 
    \frac{1 - \cos\alpha_2 e^{-R_1 \text{TR}} - \left(1 - \cos\alpha_2\right) e^{-R_1\text{TR}_{2}}}
    {1 - \cos\alpha\cos\alpha_2 e^{-R_1 \text{TR}}} ~,
\end{equation}
Here, $\alpha_2$ is unknown (it depends on the degree of magnetisation transfer saturation) and, following \citet{Helms2008b}, we name $\delta = 1 - \cos\alpha_2$ the MT saturation. Furthermore, we assume that TR$_{2} = 0$, and the signal equation simplifies to:
\begin{equation}
    S_{\text{TE}=0} = A \sin\alpha 
    \frac{(1-\delta)\left(1 - \exp\left(-R_1\text{TR}\right)\right)}
    {1 - (1-\delta)\cos\alpha\exp\left(-R_1 \text{TR}\right)}  ~.
\end{equation}

In practice, the MR environment is not perfect and multiple phenomena cause deviations from this theoretical model and introduce artefacts:
\begin{itemize}
    \item The excitation field, $B_1^+$, is not perfectly homogeneous, causing the effective flip-angle to be location-dependent and to deviate from its nominal value. Multiple techniques allow this field to be measured \citep{Jiru2006, Yarnykh2007a, Sacolick2010a}.
    \item The receive field, $B_1^-$, suffers from the same issue, especially when high-density phased-array coils are used. This is particularly problematic when between-scan motion happens, as the receive field modulation then has a different value in the different contrasts \citep{Papp2016a}.
    \item The main magnetic field, $B_0$, is not perfectly flat either, due to changes in magnetic susceptibility at air-tissue interfaces and imperfect shimming. As odd and even echoes are acquired with a readout of inverse polarity, between-echo distorsion is possible in regions where field inhomogeneity is high \citep{Varadarajan2020}. 
    \item Even with the use of gradient and RF spoiling, the assumption that no transverse magnetisation persists across TRs is rarely met and can lead to additional dependencies, on both tissue properties and hardware, in the $R_1$ estimates \citep{Preibisch2009b, Corbin2021}.
\end{itemize}

\subsection{Noise model: Gaussian}

The measured signal is hampered by thermal noise, which is uniform and Gaussian in each complex coil image and becomes Rician or non-central Chi distributed in the sum-of-square magnitude image. In the high SNR regime, this noise is approximately Gaussian, and we make that approximation in this work. This means that the fitting procedure -- which consists of inferring the unknown parameters from noisy measurements -- is effectively a non-linear least squares problem.

\subsection{Optimisation: non-linear least squares}
\label{sub:nls}

A non-linear least square problem is one of the form:
\begin{equation}
    \min_{\vec{y}} \left\{ \ell(\vec{y}) \defeq \frac{1}{2}\sum_i \left(f_i\left(\vec{y}\right) - x_i\right)^2 \right\},
    \label{eq:nls}
\end{equation}
where all $f_i$ are non-linear functions. In general, there is no closed-form solution (when there is a solution at all), and the problem can be non-convex. Therefore, carefully crafted iterative methods must be used. A method of choice is Gauss-Newton optimisation. Gauss-Newton emerges from Newton's method, that is, from a second order Taylor expansion of the objective function. Let us write the gradient of function $f$ in $\vec{y}$ as $\vec{g}_{f}(\vec{y})$ and its Hessian as $\vec{H}_{f}(\vec{y})$. The Taylor expansion of equation \eqref{eq:nls} about $\vec{y}_0$ gives:
\begin{align}
    \ell(\vec{y}) 
    & \approx \ell(\vec{y}_0) 
    + \left(\vec{y} - \vec{y}_0\right)\T\vec{g}_{\ell}(\vec{y}_0)
    \nonumber\\
    & \phantom{{}={}} 
    +  \frac{1}{2}\left(\vec{y} - \vec{y}_0\right)\T \vec{H}_{\ell}(\vec{y}_0) \left(\vec{y} - \vec{y}_0\right)
    \\
    \vec{g}_{\ell}(\vec{y}_0)
    & = \sum_i \left(f_i\left(\vec{y}_0\right) - x_i\right) \vec{g}_{f_i}(\vec{y}_0)
    \label{eq:chainrule}
    \\
    \vec{H}_{\ell}(\vec{y}_0)
    & = \sum_i \vec{g}_{f_i}(\vec{y}_0) \vec{g}_{f_i}(\vec{y}_0)\T + \left(f_i\left(\vec{y}_0\right) - x_i\right) \vec{H}_{f_i}(\vec{y}_0)
\end{align}
Under the assumption that $\vec{H}_{\ell}(\vec{y}_0)$ is positive-definite, Newton's method solves the surrogate quadratic problem by finding its stationary point:
\begin{equation}
    \vec{y}_1 = \vec{y}_0 - \vec{H}_{\ell}(\vec{y}_0)^{-1} \vec{g}_{\ell}(\vec{y}_0) ~.
\end{equation}
However, as previously stated, the problem is not necessarily convex and therefore the Hessian not necessarily positive-definite. Gauss-Newton circumvents this issue using Fisher's scoring, \emph{i.e.}, by assuming -- in the Hessian -- that residuals are all zero:
\begin{equation}
    \vec{P}_{\ell}(\vec{y}_0)
    = \sum_i \vec{g}_{f_i}(\vec{y}_0) \vec{g}_{f_i}(\vec{y}_0)\T  ~.
\end{equation}
We denote this alternative Hessian $\vec{P}$ for \emph{preconditioner}. The resulting quantity is (under mild conditions) positive-definite, and the surrogate quadratic problem now has a minimum. When it converges, the Gauss-Newton iteration converges to a stationary point of the objective function. However, convergence is not guaranteed \citep{Mascarenhas2014}. In the following subsections, we introduce a sufficient condition for monotonic convergence in Newton-like methods and show simple examples related to MR relaxometry where Gauss-Newton does not converge monotonically or even diverges. We then introduce a new preconditioner that is shown (experimentally) to enforce the condition for monotonic convergence. This preconditioner will be used to optimise the non-linear SPGR least square problem.

\subsubsection*{A sufficient condition for monotonic convergence}

Let us focus on the simpler 1D case. Let $\ell: \mathbb{R} \rightarrow \mathbb{R}$ be a twice-differentiable non-linear function over the real line with a single local (and therefore global) minimum $y^\ast$ and no local maximum\footnote{This proof could be easily extended to local-convergence when more than one minimum exists, but we focus on the simpler case for clarity}.
This function does not need to be convex but, by construction, its first derivative $g: \mathbb{R} \rightarrow \mathbb{R}$ is negative everywhere on the left of its optimum and positive everywhere on its right. Let $h: \mathbb{R} \rightarrow \mathbb{R}_\ast^+$ be a strictly positive function that we will use as a preconditioner (\emph{e.g.}, the absolute value of the curvature). Applying a Newton-like step at point $y_n$ gives an update of the form $y_{n+1} = y_n - g(y_n)/h(y_n)$. Since $h$ is positive, the step is ensured to be in the direction of the optimum; a sufficient condition for this step to improve the objective function ($\ell(y_{n+1}) < \ell(y_n)$) is for the updated point to fall \emph{between} $y_n$ and the optimum $y^\ast$. Finally, this condition can be formalised as
\begin{equation}
    \frac{\left|g(y_n)\right|}{h(y_n)} \leqslant \left| y_n - y^\ast \right| ~.
    \label{eq:cond_monotonic}
\end{equation}
While this condition seems difficult to enforce or check, note that it is always true at the optimum, where $g(y^\ast) = 0$. Let us write $r(y) = g(y)/h(y)$ and add the condition that $h$ is differentiable; then condition \eqref{eq:cond_monotonic} is enforced if the (absolute value of the) derivative of $r$ is everywhere equal or below 1. On the right (positive) side, $r'(y) \leqslant 1$ on $[y^\ast, y_n]$ implies
$$
|r(y_n)| = r(y_n) = \int_{y^\ast}^{y_n} r'(y) \diff y \leqslant \int_{y^\ast}^{y_n} 1 \diff y = y_n - y^\ast = |y_n - y^\ast| ~. 
$$
The same thing can be shown on the left (negative) side.

In the multivariate case, we want the update to stay in the same ``quadrant'' (that is, all coordinates fall between their original value and the optimum), which we formalise as:
\begin{equation}
    \left|\vec{H}(\vec{y}_n)^{-1}\vec{g}(\vec{y}_n)\right| \leqslant \left| \vec{y}_n -\vec{y}^\ast \right| ~,
\end{equation}
where $\left| ~\cdot~ \right|$ is the element-wise absolute value. 
However, we can apply any arbitrary rotation to the problem and this condition stays true. We can therefore relax it to:
\begin{equation}
    \lVert\vec{H}(\vec{y}_n)^{-1}\vec{g}(\vec{y}_n)\rVert \leqslant \lVert \vec{y}_n - \vec{y}^\ast \rVert ~,
\end{equation}
where $\lVert ~\cdot~ \rVert$ is the Euclidean norm.

\subsubsection*{Toy problem 1: exponential least square}

\begin{figure*}[t!]
    \centering
    \includegraphics[width=\textwidth]{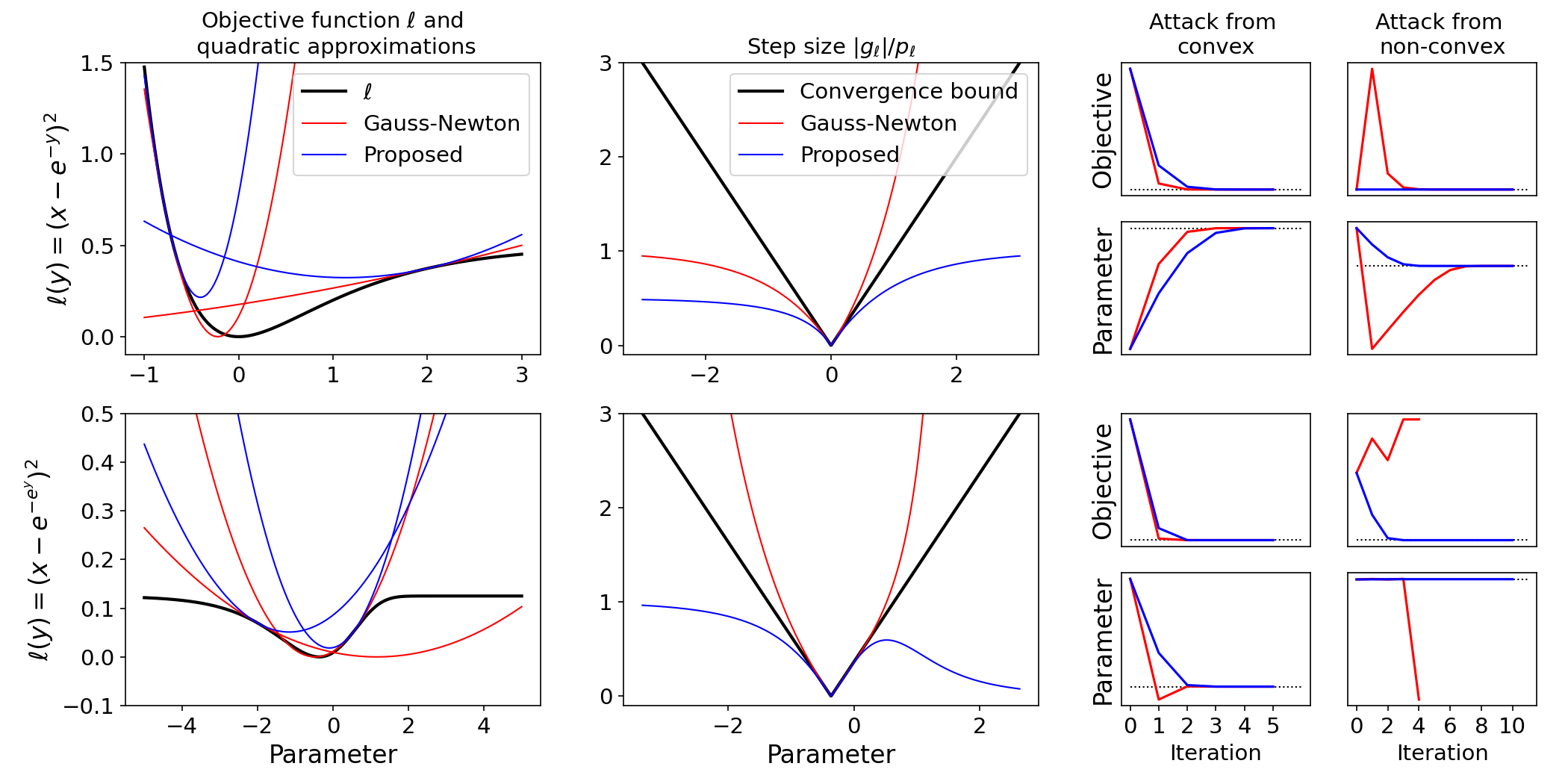}
    \caption{%
    Qualitative analysis of two toy problems related to MR relaxometry. The first toy problem (first row) has an objective function of the form $\left(x - \exp(-y)\right)^2$, while the other toy problem (second row) as one of the form $\left(x - \exp(-\exp(y))\right)^2$. 
    In each case, the first column shows the objective function in black along with quadratic approximations corresponding to Gauss-Newton (red) and the proposed second order method (blue). The second column shows the magnitude of the gradient-hessian ratio. The black line corresponds to a purely quadratic objective function, whose optimum is found in one step ($|g(x)|/h(x) = \left|x-x^\ast\right|$). When the ratio is under this black line, monotonic convergence is ensured. The third column shows convergence of the Gauss-Newton (red) and proposed (blue) iterative schemes when attacking from either a convex or non-convex lobe, with the optimum shown in dashed black. In each case, convergence is shown both in terms of objective value per iteration and parameter value per iteration.%
    }
    \label{fig:nls}
\end{figure*}

Let us now discuss two toy problems that are closely related to MR relaxometry. The first one consists of finding the least square solution when the non-linear function is an exponential and there is a single data point:
\begin{equation}
    \ell(y) = \frac{1}{2} \left(e^{-y} - x\right)^2
\end{equation}
The corresponding energy landscape, shown as a solid black line in Fig. \ref{fig:nls}, is non-convex on its left. When $x > 0$, there is an obvious analytical solution ($y^\ast = -\log(x)$), but we will use an iterative method to get to the same conclusion. Using the same notations as before, we have:
\begin{align}
    g_\ell(y) & = -(e^{-y} - x) ~e^{-y} ~,\\
    h_\ell(y) & = e^{-2y} + (e^{-y} - x) ~e^{-y} ~,\\
    p_\ell(y) & = e^{-2y} ~.
\end{align}
If we compute the derivative of $r_\ell(y) = g_\ell(y)/p_\ell(y)$, we find
\begin{equation}
    r'_\ell(y) = -x e^{y} ~,
\end{equation}
whose absolute value is \emph{not} bounded by 1. We therefore introduce an alternative pre-conditioner that uses the absolute value of the second term rather than discarding it:
\begin{align}
    \tilde{p}_\ell(y) & = e^{-2y} +  \left|e^{-y} - x\right|~e^{-y}  ~.
\end{align}
Although we do not prove it, we find experimentally that this preconditioner enforces condition \eqref{eq:cond_monotonic}. The corresponding quadratic approximations are plotted in red (Gauss-Newton) and blue (proposed preconditioner) in Fig. \ref{fig:nls}. Experimental convergence with both preconditioners is also shown. Conversely to Gauss-Newton, the proposed version is monotonic when attacking from the non-convex side and converges almost as fast as Gauss-Newton when attacking from the convex side. 

\subsubsection*{Toy problem 2: nested exponential least square}

Our second toy problem involves nested exponentials:
\begin{equation}
    \ell(y) = \frac{1}{2} \left(x - e^{-e^{y}}\right)^2 .
\end{equation}
The corresponding energy landscape is shown in Fig. \ref{fig:nls}: its curvature now tends towards zero on both sides of the real line. Its gradient, Hessian and preconditioners are:
\begin{align}
    g_\ell(y_0) & = -e^{y_0-e^{y_0}} ~(e^{-e^{y_0}} - x) ~,\\
    h_\ell(y_0) & = \left(e^{y_0-e^{y_0}}\right)^2 + (e^{2y_0} - e^{y_0})~e^{-e^{y_0}}~(e^{y_0} - x) ~,\\
    p_\ell(y_0) & = \left(e^{y_0-e^{y_0}}\right)^2 ~, \\
    \tilde{p}_\ell(y_0) & = \left(e^{y_0-e^{y_0}}\right)^2 + \left| (e^{2y_0} - e^{y_0})~e^{-e^{y_0}}~(e^{y_0} - x) \right| ~.
\end{align}
Again, $\left|r_\ell'\right|$ is not bounded, and the Gauss-Newton preconditioner fails to enforce the convergence condition. Experimental convergence is shown in Fig. \ref{fig:nls}, where the proposed method is shown to monotonically converge while Gauss-Newton can diverge when starting in a non-convex section of the energy landscape.

\subsubsection*{Generalisation to the multivariate case}

\begin{figure}
    \centering
    \includegraphics[width=\columnwidth]{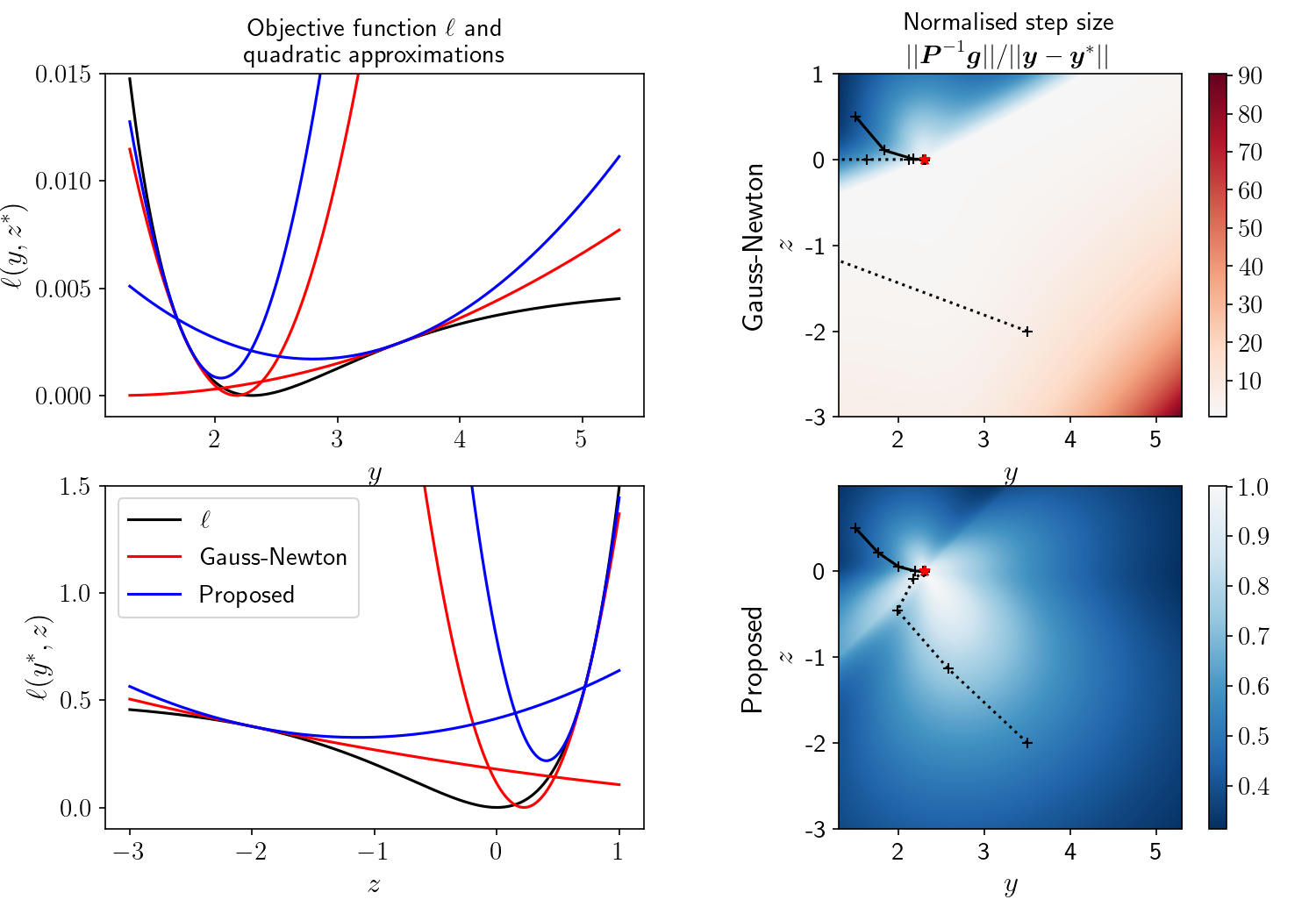}
    \includegraphics[width=\columnwidth]{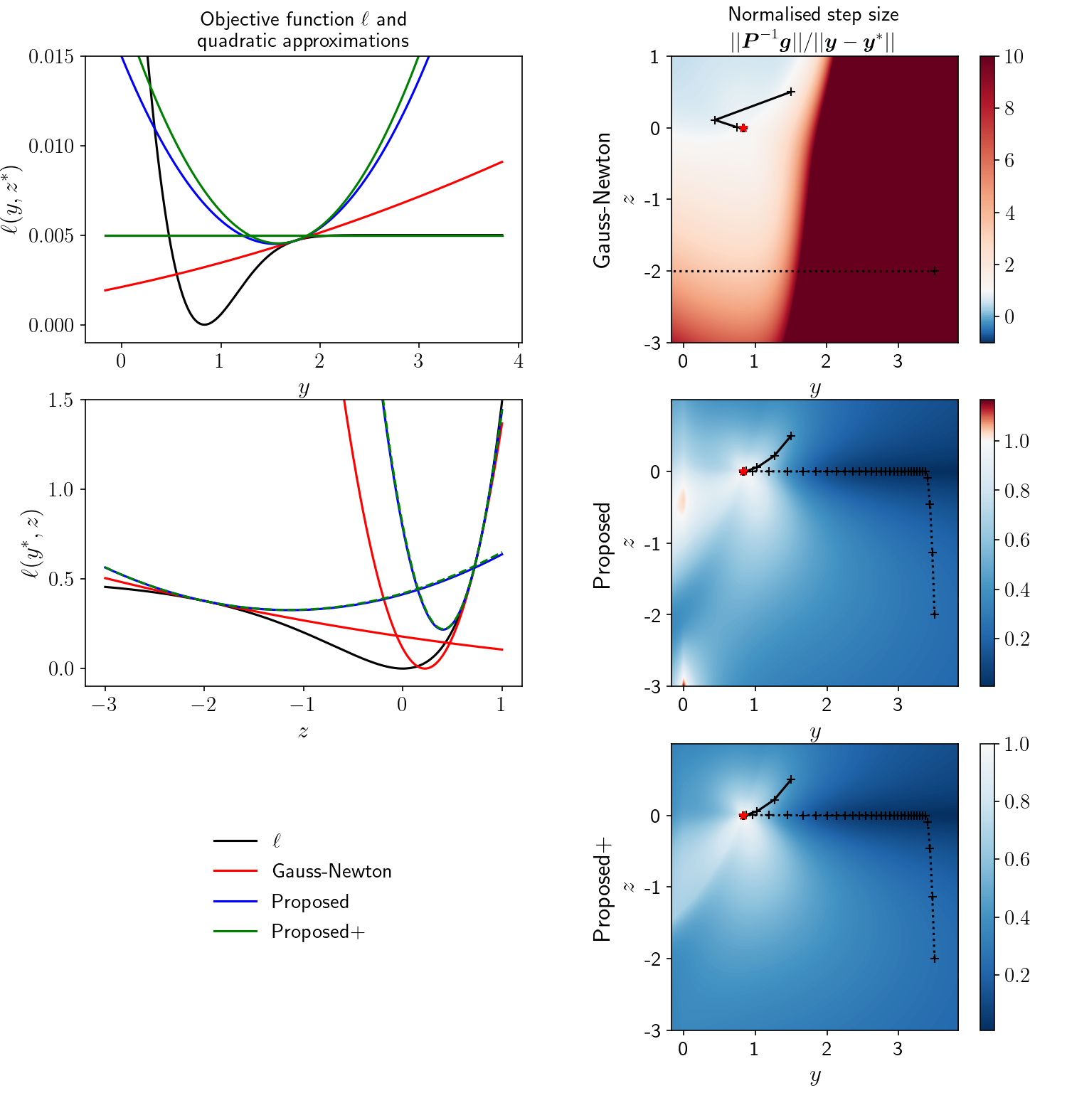}
    \caption{%
    Qualitative analysis of two-dimensional toy problems related to MR relaxometry. The first toy problem (top half) has an objective function of the form $\ell(y, z) = \left(x_0 - e^{z-t_0y}\right)^2 + \left(x_1 - e^{z-t_1 y}\right)^2$, while the other toy problem (bottom half) has one of the form $\ell(y, z) = \left(x_0 - e^{z-t_0e^y}\right)^2 + \left(x_1 - e^{z-t_1 e^y}\right)^2$. 
    In each case, the first column shows the objective function along each of the two dimensions (in black) along with quadratic approximations corresponding to Gauss-Newton (red) and the proposed second order methods (blue and green). The second column shows the normalised step size for each method, color-coded such that values below one are blue and values above one are red. The optimum is marked by a red star. Convergence paths from two random initial coordinates (solid and dashed curves) are overlaid. In the second toy problem, two different loading matrices are used: \emph{Proposed} uses the diagonal of the absolute Hessian $\diag(\left|\vec{H}\right|)$ whereas \emph{Proposed+} uses its sum across columns $\left|\vec{H}\right|\vec{1}$.%
    }
    \label{fig:nls2d}
\end{figure}

The toy problems investigated in the previous section are highly relevant to MR relaxometry: the first case is related to the exponential fit of the signal decay, where each data point is seen as the exponential of a linear combination of parameters, while the second case is an extension of the first, where positive parameters are encoded by their log. Given the apparent robustness of the proposed preconditioner on these toy problems, we propose two extensions to the general multivariate case by replacing the absolute value of the scalar Hessian in the one-dimensional case with either: (1) the absolute value of the diagonal of the Hessian, $\left|\vec{H}_{f_i}(\vec{y}_0)\right| \odot \vec{I}$; (2) the sum of the absolute value of the Hessian across columns, $\diag\left(\left|\vec{H}_{f_i}(\vec{y}_0)\right| \vec{1}\right)$. Only the second one is ensured to majorise the true Hessian \citep{Chun2018}, but we show in our experiments that the first one yields a stable convergence in the MPM context. We therefore use the preconditioner:
\begin{align}
    \textstyle
    \tilde{\vec{P}}_{\ell}(\vec{y}_0)
    = \sum_i & \vec{g}_{f_i}(\vec{y}_0) \vec{g}_{f_i}(\vec{y}_0)\T 
    + \left| f_i\left(\vec{y}_0\right) - \vec{x}_i\right|~ \left|\vec{H}_{f_i}(\vec{y}_0)\right| \odot \vec{I}.
    \label{eq:precond}
\end{align}
The impact of this loading is illustrated with a pair of two-dimensional toy problems that are closely related to the one-dimensional problems presented before. The first problem involves an objective function of the form $\ell(y, z) = \frac{1}{2}\left(x_0 - e^{z-t_0y}\right)^2 + \frac{1}{2}\left(x_1 - e^{z-t_1 y}\right)^2$. In the second problem, the decay parameter is encoded by its log: $\ell(y, z) = \frac{1}{2}\left(x_0 - e^{z-t_0e^y}\right)^2 + \frac{1}{2}\left(x_1 - e^{z-t_1 e^y}\right)^2$. Step sizes and optimisation trajectories are depicted in Fig. \ref{fig:nls2d}. When the Gauss-Newton preconditioner is used, a vast part of the parameter space has very large step sizes that lead to overshooting. When the proposed preconditioners are used, the entire parameter space (or most of it) lies below the monotonic convergence threshold.

\subsection{Likelihood: spoiled gradient echo}

\begin{figure}[t!]
    \centering
    \includegraphics[width=\columnwidth]{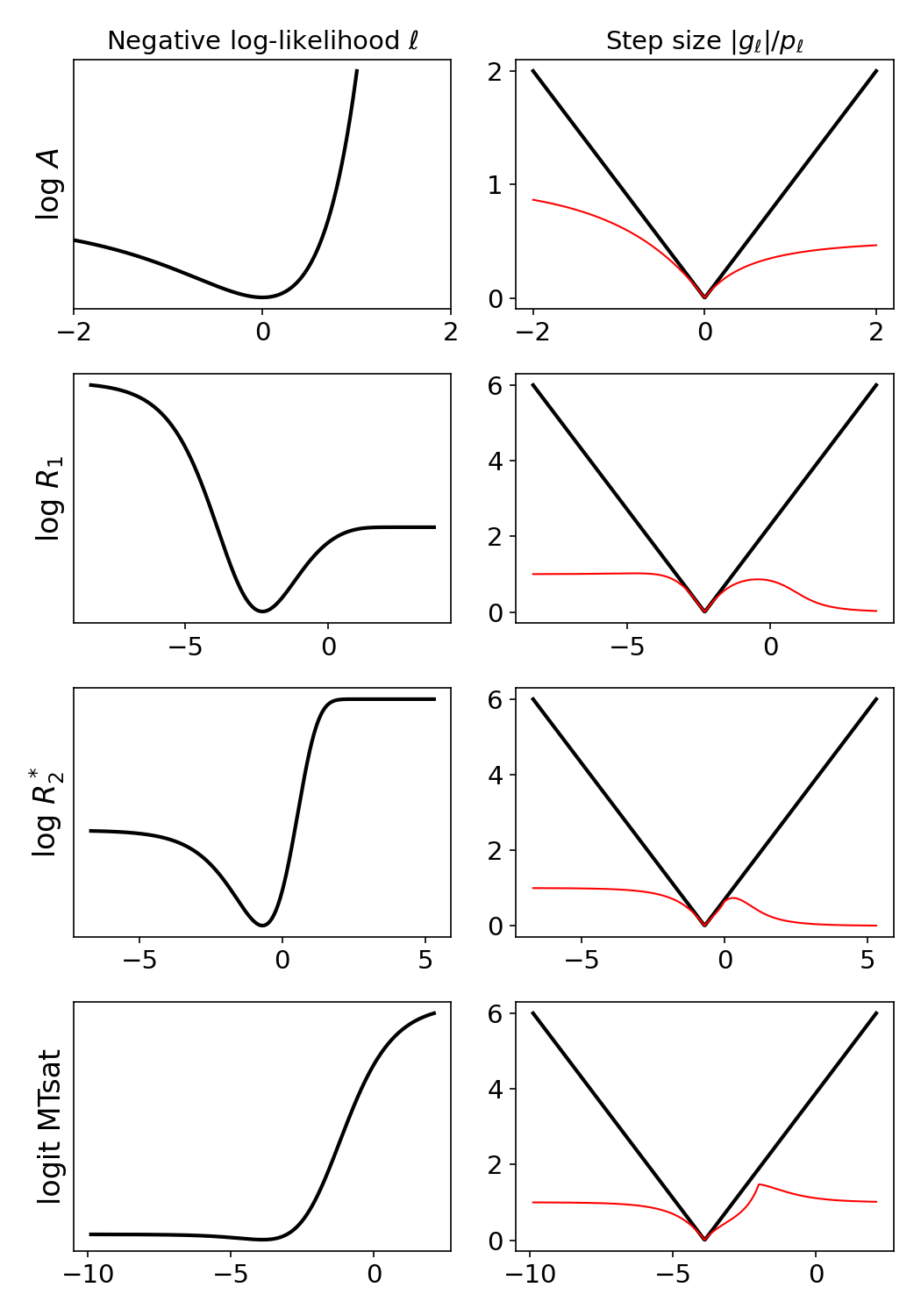}
    \caption{Negative log-likelihood and step size function of the nonlinear SPGR model. The first column shows the log-likelihood of a single observation with respect to each log-parameter, while keeping the others fixed. The second column shows the step size function (red) obtained with the proposed approximate Hessian, compared with the theoretical perfect step size function $f(x) = |x - x^\ast|$. As long as the actual step size is below this theoretical function, monotonic convergence is ensured.}
    \label{fig:gre_conv}
\end{figure}

The main contribution of this paper lies in the direct optimisation of $A$, $R_1$ $R_2^\ast$ and $\delta$. First, each parameter is encoded by its log (or logit in the case of $\delta$). Let us switch from a physics to a mathematical notation, where scalars and vectors are written in lowercase while capital letters are reserved for matrices and tensors. We use a tilde to differentiate the log-parameters from their exponentiated version, so that in each voxel:
\begin{alignat*}{4}
    & a && = \exp\left(\tilde{a}\right), ~~~~~&& r_1 && = \exp\left(\tilde{r}_1\right), \\
    & r_2 && = \exp\left(\tilde{r}_2\right), ~~~~~&& \delta && = \frac{1}{1 + \exp\left(-\tilde{\delta}\right)} ~.
\end{alignat*}
Let us assume that all images are defined on the same grid with $N$ voxels. We write the set of log-parameters as $\vec{Y} = \left[\tilde{\vec{a}}, \tilde{\vec{r}}_1, \tilde{\vec{r}}_2, \tilde{\vec{\delta}}\right] \in \mathbb{R}^{N\times4}$ and the set of observed 3D images as $\vec{X} \in \mathbb{R}^{N \times I}$. Observed images come with a set of fixed parameters $\vec{\Theta} \in \mathbb{R}^{4 \times I}$, where each quadruplet $\vec{\theta}_i = \left(\alpha_i, t_{ri}, t_{ei}, \sigma_i^2\right) \in \mathbb{R}^4$ contains the flip angle, repetition time, echo time and noise variance of an image. The likelihood term of the objective function involves a sum across images followed by a sum across voxels:
\begin{equation}
    \ell(\vec{Y}) = \sum_i \sum_n \frac{1}{2\sigma_i^2} \left(x_{in} - s(\vec{y}_n \mid \vec{\theta}_i)\right)^2 ~,
\end{equation}
where $s$ is the intensity of a spoiled gradient echo\footnote{%
When the receive and transmit fields 
are known from \emph{e.g.} field mapping, equation \eqref{eq:spgr_signal} is slighlty changed. The flip angle $\alpha$ takes a voxel-dependent value $\alpha_n = \alpha_{\text{nom}} b_n^+$, where the nominal flip angle is modulated by the transmit field efficiency; and the whole signal is modulated voxel-wise by the net receive field $b_n^-$.}:
\begin{equation}
    s(\tilde{a}, \tilde{r}_1, \tilde{r}_2, \tilde{\delta} \mid \alpha, t_r, t_e) = 
    a\sin\alpha
    \frac{(1-\delta)(1-e^{-r_1 t_r})}{1 - (1-\delta)\cos\alpha e^{-r_1 t_r}}
    e^{-r_2t_e} ~.
    \label{eq:spgr_signal}
\end{equation}
For contrasts that do not include an MT pulse, $\delta = 0$ and gradients with respect to $\tilde{\delta}$ are zero. We recognise a non-linear least squares problem, and propose to use the preconditioner from equation \eqref{eq:precond}, which depends on the gradient and Hessian of $s$. Let us focus on a single voxel and drop the $(i, n)$ subscripts for clarity. After differentiation, the gradient is:
\begin{align}
\frac{\partial s}{\partial \tilde{a}} & = s ~,\\
    \frac{\partial s}{\partial \tilde{r}_1} & = -t_r r_1 ~
    \frac{((1-\delta)\cos\alpha - 1)~ e^{-r_1 t_r}}
    {(1 - (1-\delta)\cos\alpha ~e^{-r_1 t_r})~(1 - e^{-r_1 t_r})} 
    ~s ~,\\
    \frac{\partial s}{\partial \tilde{r}_2} & = -t_e r_2 ~s ~,\\
    \frac{\partial s}{\partial \tilde{\delta}} & = \frac{\delta}{(1-\delta)\cos\alpha~e^{-r_1 t_r}-1} ~s ~,
\end{align}
and the Hessian is:
\begin{align}
\frac{\partial^2 s}{\partial \tilde{a}^2} & = \frac{\partial s}{\partial \tilde{a}} ~,\\
    \frac{\partial^2 s}{\partial \tilde{r}_1^2} & = \left(1 - t_r r_1 ~ \frac{1 + (1-\delta)\cos\alpha~e^{-r_1 t_r}}{1 - (1-\delta)\cos\alpha~e^{-r_1 t_r}}\right) ~\frac{\partial s}{\partial \tilde{r}_1} ~,\\
    \frac{\partial^2 s}{\partial \tilde{r}_2^2} & = \left(1 - t_e r_2\right) ~\frac{\partial s}{\partial \tilde{r}_2} ~,\\
    \frac{\partial^2 s}{\partial \tilde{\delta}^2} & = 
    \left(\frac{2(1-\delta)(1-\cos\alpha~e^{-r_1 t_r})}{1 - (1-\delta)\cos\alpha~e^{-r_1 t_r}} - 1\right)
    \frac{\partial s}{\partial \tilde{\delta}} ~.
\end{align}
The gradient $\vec{g}_\ell$ and preconditioner $\tilde{\vec{P}}_\ell$ of the least-square objective function are then computed as in equations \eqref{eq:chainrule} and \eqref{eq:precond} and summed across echo times and contrasts. The log-likelihood of one observation with respect to each parameter -- keeping the others fixed -- is shown in Fig. \ref{fig:gre_conv}, along with the step size function, which is consistently under the monotonic convergence bound. Although this figure only shows the step size function for one given set of parameters, the convergence property is consistently enforced when sampling random parameters and observed values.

\subsubsection*{Spatial projection}

In practice, it is common for people to move their head between scans, making inter-scan registration a mandatory initial step. While reslicing all scans in the same space is a possibility, we prefer to keep the raw data untouched and integrate the spatial projection from the reconstruction space to the different acquisition spaces in the forward model, modifying slightly the objective function. Let us write all log parameter maps as $\vec{y} = \operatorname{vec}(\vec{Y}) \in \mathbb{R}^{NK}$, where $N$ is the number of voxels in reconstruction space and $K$ the number of maps, and $\vec{\Psi} \in \mathbb{R}^{MK\times NK}$ a matrix that encodes linear reslicing from reconstruction space to one acquisition space\footnote{Since the same transformation is applied to all maps, $\vec{\Psi}$ really is the Kronecker product of an $M\times N$ reslicing matrix and a $K \times K$ identity matrix.}. The new objective function is $\ell_\psi\left(\vec{y}\right) = \ell\left(\vec{\Psi}\vec{y}\right)$. Application of the chain rule yields:
\begin{align}
    \frac{\partial\ell_\psi}{\partial\vec{y}} & = \vec{\Psi}\T\left(\frac{\partial\ell}{\partial\vec{y}}\left(\vec{\Psi}\vec{y}\right)\right) ~, \\
    \frac{\partial^2\ell_\psi}{\partial\vec{y}\partial\vec{y}\T} & = \vec{\Psi}\T\left(\frac{\partial^2\ell}{\partial\vec{y}\partial\vec{y}\T}\left(\vec{\Psi}\vec{y}\right)\right)\vec{\Psi} ~.
\end{align}
However, this new Hessian is much less sparse: it cannot be seen as block diagonal anymore. It was shown in \citet{Ashburner2018} that this matrix can be majorised (in the L\"owner order sense) by a block-diagonal matrix:
\begin{align}
    \frac{\partial^2\ell_\psi}{\partial\vec{y}\partial\vec{y}\T} \preceq
    \diag\left(\vec{\Psi}\T\left(\frac{\partial^2\ell}{\partial\vec{y}\partial\vec{y}\T}\left(\vec{\Psi}\vec{y}\right)\right)\right)
    ~.
\end{align}
In other words, for each contrast, the log parameter maps are resliced (\emph{pulled}) to the acquisition space, where the gradient and Hessian of the least square function are computed and sent back to recon space using the adjoint of the reslicing operation (\emph{pushed}).

\subsection{Prior: joint total variation}

Multiple types of spatial regularisation exist, with different properties. Here, we focus on joint total-variation, equivalent to an $\ell_{1,2}$ norm over spatial gradients, which benefits from the same sparsity-inducing properties as other $\ell_{1}$ norms, except that sparsification is shared across contrasts. This property is particularly interesting when the same object is seen through different contrasts, as is the case in the context of MPMs, since edges are shared across parameter maps. Again, let $\vec{Y} = \left[\tilde{\vec{a}}, \tilde{\vec{r}}_1, \tilde{\vec{r}}_2, \tilde{\vec{\delta}}\right] \in \mathbb{R}^{N\times4}$ be the set of all parameter maps, the regulariser has the form:
\begin{equation}
    \operatorname{JTV}(\vec{Y}) = \sum_{n=1}^N \sqrt{\sum_{k=1}^4 \lambda_k \left(\vec{G}_n\vec{y}_{k}\right)\T\left(\vec{G}_n\vec{y}_{k}\right)} ~,
\end{equation}
where $\vec{G}_n$ extracts all 6 forward and backward finite-differences (in 3D) about the $n$-th voxel and $\lambda_k$ is a parameter-specific regularisation factor. 

This regulariser is convex but non-smooth, complicating its use in optimisation. However, several methods have been developed over the years to solve this problem; \emph{e.g.}, ISTA \citep{Daubechies2004}, FISTA \citep{Beck2009}, ADMM \citep{Boyd2011} and variations thereof. In this paper, we use an iteratively reweighted least squares (IRLS) scheme \citep{Daubechies2010, Bach2011}, which relies on the bound:
\begin{multline}
    {\textstyle
    \sqrt{\sum_k \lambda_k \left(\vec{G}_n\vec{y}_{k}\right)\T\left(\vec{G}_n\vec{y}_{k}\right)}}
    \\ = \min_{w_n > 0}
    {\textstyle \left\{ \frac{1}{2w_n} + \frac{w_n}{2} \sum_k \lambda_k \left(\vec{G}_n\vec{y}_{k}\right)\T\left(\vec{G}_n\vec{y}_{k}\right) \right\} .}
    \label{eq:bound}
\end{multline}
When the weight map $\vec{w}$ is fixed, this bound can be seen as a Tikhonov (\emph{i.e.}, $\ell_2$) prior with nonstationary regularisation, which is a quadratic prior that factorises across parameters. Therefore, the between-parameters correlations induced by the JTV prior are entirely captured by the weights. Conversely, when the parameter maps are fixed, the weights can be updated in closed-form:
\begin{align}
    \textstyle
    w_n = \left(\sum_k \lambda_k \left(\vec{G}_n\vec{y}_{k}\right)\T\left(\vec{G}_n\vec{y}_{k}\right)\right)^{-\frac{1}{2}} ~.
\end{align}

Let us now vectorise all log-parameter maps: $\vec{y} = \operatorname{vec}\left(\vec{Y}\right)$. The quadratic term in equation \eqref{eq:bound} -- when summed over all voxels -- can be written as $\vec{y}\T\vec{L}\vec{y}$ with $\vec{L}=\vec{G}\T\vec{W}\vec{G} \otimes \diag\left(\vec{\lambda}\right)$, where $\vec{G}\in\mathbb{R}^{6N\times N}$ extracts all 6 forward and backward finite-differences about all voxels and $\vec{W} = \diag(\vec{w}) \otimes \vec{I}_6$. Note further that $\vec{G}\T\vec{G}$ is Toeplitz and can be implemented as a convolution by a small kernel \citep{Ashburner2007}. Similarly, $\diag\left(\vec{G}\T\vec{W}\vec{G}\right)$ can be obtained by convolving the weight map $\vec{w}$ with a small ``cross'' kernel.

\subsection{Optimisation: regularised spoiled gradient echo}

Optimisation follows an IRLS scheme, where each iteration consists of alternately solving for the parameter maps $\vec{y}$ and the JTV weights $\vec{w}$. At each iteration, the objective function in $\vec{y}$ is of the form:
\begin{equation}
    \ell(\vec{y}) = \sum_{c,e} \frac{1}{2\sigma_c^2} \lVert \vec{x}_{ce} - s_{ce}\left(\vec{y}\right) \rVert^2 + \frac{1}{2}\vec{y}\T\vec{L}\vec{y} ~,
\end{equation}
where $c$ and $e$ are contrasts and echoes. This function is non-linear and solved using the proposed second order method, from which we get a gradient $\vec{g} \in \mathbb{R}^{4N}$ and an approximate Hessian $\vec{H} \in \mathbb{R}^{4N \times 4N}$. This matrix is actually block diagonal as it consists of $N$ small $4 \times 4$ matrices. Performing the Newton update step involves solving the linear system
\begin{equation}
    \left(\vec{H} + \vec{L}\right)^{-1}\left(\vec{g} + \vec{L}\vec{y}\right) ~,
\end{equation}
which is done using the preconditioned conjugate gradient method. The preconditioner that we use is 
$$\vec{H} + \left(\vec{G}\T\vec{W}\vec{G}\odot \vec{I}\right)\otimes\diag\left(\vec{\lambda}\right) ~,$$

\subsection{Implementation}

In practice, 10 IRLS iterations were used, with 5 inner Newton iterations and 32 conjugate gradient iterations per Newton step. A tolerance threshold on the gain between two subsequent iterations was set for early stopping (IRLS: $10^{-5}$, Newton: $10^{-5}$, CG: $10^{-3}$). The variance of the thermal noise was estimated by fitting a two-class Rician mixture to each acquired volume \citep{Ashburner2013}; the class with lowest mean was considered as the background. Variances were then averaged across echo times using the geometric mean. Initial (constant) parameter maps were estimated using the mean of the other (non-background) class in each volume.

This algorithm, along with baseline methods, is implemented in NiTorch, a Python library with a PyTorch backend dedicated to neuroimaging. It is publicly available at \url{https://github.com/balbasty/nitorch}. The processing of a 0.8 mm MPM dataset -- from which $A$, $R_1$, $R_2^\ast$ and MT\textsubscript{sat} can be estimated -- fits in 12GB of memory and takes about 10 minutes on a Nvidia Titan V GPU for SPGR+JTV (5 minutes for NONLIN+JTV, $<$ 1 minute for LOGLIN).

\section{Experiments}
\label{sec:experiments}

\subsection{Convergence analysis}

This experiment was based on simulated values. One thousand single-voxel datasets were simulated, each with their own relaxation parameters, TRs, TEs and flip angles. Each voxel had three contrasts (i.e., three different TRs and flip angles) -- including one with an MT pulse -- and five echo times per contrast. All log parameters ($\log A$, $\log R_1$, $\log R_2^\ast$, logit MT\textsubscript{sat}, log TR, log TE) were uniformly sampled in $[-5, 5]$, while flip angles were uniformly sampled in $[0, \frac{\pi}{4}]$ and the noise variance was fixed at $\sigma^2 = 1$. The (unregularised) SPGR model was then fitted using 10,000 iterations of the proposed second-order optimisation, and the negative log-likelihood was computed after each iteration. Two baseline algorithms were used as well: Gauss-Newton with a backtracking line-search (at each step, the descent direction was modulated by an Armijo factor) and Levenberg-Marquardt (LM). Following \cite{Press2007}, updates were only accepted if they improved the objective-function. On failure, the GN Armijo factor (respectively the LM damping parameter) was divided (resp. multiplied) by 10, while it was multiplied (resp. divided) by 10 on success. One Armijo or damping factor was defined in each virtual voxel.

A similar experiment was conducted in order to compare the use of a log, rate or time representation during optimisation. Two alternative optimisation schemes were derived such that either ($A$, $R_1$, $R_2^\ast$, MT$_\text{sat}$) or ($A$, $T_1$, $T_2^\ast$, MT$_\text{sat}$) are optimised instead of ($\log A$, $\log R_1$, $\log R_2^\ast$, $\operatorname{logit}\text{MT}_\text{sat}$). These two representations required an even more stable Hessian, were $\left|\vec{H}\right| \vec{1}$ is used to load the diagonal instead of  $\diag\left(\left|\vec{H}\right|\right)$.

\subsection{Datasets}

Two dataset were used in this study:
\begin{enumerate}
    \item The first dataset used in this study is a single-subject, single-session reproducibility dataset: a single participant was scanned five times in a single session with a 0.8 mm MPM protocol. This protocol consists of three multi-echo gradient echo contrasts (flip angle: 21/6/6 deg, number of echoes: 8/8/6 with 2.3 ms echo spacing, off-resonance pulse: No/No/Yes, TR: 25 ms, resolution: 0.8 mm, FOV: $256 \times 224 \times 179$ mm$^3$, GRAPPA acceleration: $2 \times 2$ with 40 reference lines). Calibration data were acquired to additionally account for contrast-specific receive field \citep{Papp2016a} and participant-specific transmit field inohomogeneities \citep{Lutti2010a} using the protocols described in \citet{Callaghan2019}. Signal-to-noise ratios were estimated by fitting a two-class Rician mixture model to the image histograms. SNRs in the first echo were 29 (PDw), 25 (T1w) and 28 (MTw). SNRs in the last echo were 16 (PDw), 15 (T1w) and 16 (MTw).
    
    \item The second dataset is the demonstration dataset from the hMRI toolbox \citep{Callaghan2019}, which consists of the same 0.8 mm MPM protocol acquired in a single subject, with motion between the MT-weighted scan and the PD- and T1-weighted scans.
    
\end{enumerate}

\subsection{Algorithms}

We compared two variants of our method (ML and MAP) to methods based on a two-step process described in following \citet{Tabelow2019} (ESTATICS to estimate $R_2^\ast$ and $T_E=0$ intercepts, followed by an analytical fit of $R_1$, $A$ and MT\textsubscript{sat} from these intercepts). These baseline methods use either a log-linear fit or a non-linear fit of ESTATICS, and use different denoising methods (pre-processing with a non-local means filter or MAP with a JTV prior):
\begin{itemize}
    \item LOGLIN performs a log-linear fit of $R_2^\ast$ from which weighted images extrapolated to TE=0 are obtained. This is equivalent to ESTATICS \citep{Weiskopf2013a}, except that the spatial projection is integrated in the model. The Hessian matrix that is used is a (slight) majoriser of the true Hessian, so three Newton-like iterations are needed to reach the optimum (compared to a single one, had the true Hessian been used). Since all TRs are identical, $A$, $R_1$ and MT\textsubscript{sat} are then computed analytically according to \cite{Mohammadi2017}, without requiring rational approximations.
    \item LOGLIN+AONLM applies an anisotropic non-local mean filter \citep{Manjon2010d} to all individual echoes before processing them with LOGLIN.
    \item NONLIN performs a non-linear fit or $R_2^\ast$ instead of a log-linear fit, but does not use any regularisation. This method was not used in \citet{Balbastre2020}, because Gauss-Newton was not stable enough without regularisation. The new Hessian used in this work makes its optimisation possible.
    \item NONLIN+JTV adds JTV-regularisation to NONLIN \citep{Balbastre2020}.
    \item SPGR-ML performs a non-linear maximum-likelihood fit of the log-parameters $\log R_1$, $\log R_2^\ast$, $\log A$ and logit MT\textsubscript{sat}.
    \item SPGR+JTV adds JTV regularisation to SPGR-ML, and effectively performs a non-linear maximum \emph{a posteriori} fit.
\end{itemize}

For each dataset, as a preprocessing step, all contrasts across all repeats (and their $B_1^{+/-}$ maps) were registered towards the PD-weighted series of the first repeat with SPM12. This co-registration step only modified the orientation matrices in the header of the NIfTI files and did not involve any reslicing.

\subsection{Cross-validation}

This experiment was based on the single-session dataset. First, one of the repeats was used to estimate optimal regularisation factors for LOGLIN+AONLM, NONLIN+JTV, and SPGR+JTV by cross-validation across echoes. Five folds were constructed, with the first six echo times randomly permuted and split between a training set -- used to estimate parameter maps -- and a testing set -- used to evaluate predictions. In each fold, the training echoes were used to estimate parameters with multiple regularisation values; these parameters were then used to simulate the left-out echoes and compared against the true (acquired) volumes. The mean-squared error (MSE) between the true and predicted echoes was computed and z-normalised across folds, contrast and echo times. Finally, the median z-normalised MSE was used as a metric to select the most adequate regularisation factor.

The remaining four repeats were used to compare all (optimised) methods by cross-validation across repeats. Parameter maps were computed for each repeat and used to predict individual echoes from all other repeats. 




\subsection{Echo decimation}

This experiment was based on the reproducibility dataset. First, reference parameter maps were obtained by computing the maximum-likelihood solution (SPGR-ML) of combined data from  all four test runs. Parameter maps of one of the runs were then reconstructed by either SPGR+JTV or SPGR-ML using the first 2, 4 or 6 echoes. The root mean squared error (RMSE) -- within a brain ROI -- was computed between each map and its reference.

\subsection{Uncertainty mapping}

The inverse of the Hessian of the objective function at the optimum can be used, under the Laplace approximation, to estimate the posterior uncertainty of the log-parameters \citep{MacKay2003}. Since it is easier to invert, we rather use the inverse of the block-diagonal preconditioner and extract its diagonal: 
\begin{equation}
    \operatorname{cov}\left[\vec{y} ~\middle|~ \mathcal{X}\right] \approx 
    \left(\vec{H} + \diag\left(\vec{G}\T\vec{W}\vec{G}\right)\otimes\vec{\lambda}\right)^{-1} ~,
\end{equation}
where $\mathcal{X}$ contains all observed volumes used to estimate the maps. Note that under the Laplace approximation, the posterior of the log-parameter maps is Normal-distributed, making the parameters themselves Log-Normal-distributed. If the parameter maps themselves are of interest, it might be more interesting to return their expected value and variance, which in the case of $R_1$ gives:
\begin{align}
    \mathbb{E}\left[r_1\right] 
    & = \sqrt{\exp\left(\tilde{\sigma}^2\right)}~\exp\left( \tilde{r}_1\right) ~,\\
    \mathbb{V}\left[r_1\right] 
    & = \left(\exp\left(\tilde{\sigma}^2\right)-1\right)\exp\left(\tilde{\sigma}^2\right)\exp\left(\tilde{r}_1\right)^2 ~.
\end{align}
Conversely, if $T_1=1/R_1$ is of interest:
\begin{align}
    \mathbb{E}\left[t_1\right] 
    & = \sqrt{\exp\left(\tilde{\sigma}^2\right)}~\exp\left( -\tilde{r}_1\right) ~,\\
    \mathbb{V}\left[t_1\right] 
    & = \left(\exp\left(\tilde{\sigma}^2\right)-1\right)\exp\left(\tilde{\sigma}^2\right)\exp\left(-\tilde{r}_1\right)^2 ~.
\end{align}
In this experiment, we used three echoes from the hMRI demonstration dataset to reconstruct maps using SPGR+JTV and SPGR-ML and computed the posterior uncertainty according to the above formula. This allowed us to map the posterior mean and standard deviation of log-$R_1$, $R_1$ and $T_1$. The ratio between the expected value of $R_1$ under the Laplace approximation and under a point estimate is given by $\sqrt{\exp\left(\tilde{\sigma}^2\right)}$.

\section{Results}
\label{sec:results}

\subsection{Convergence}

\begin{figure}[t!]
    \centering
    \includegraphics[width=\columnwidth]{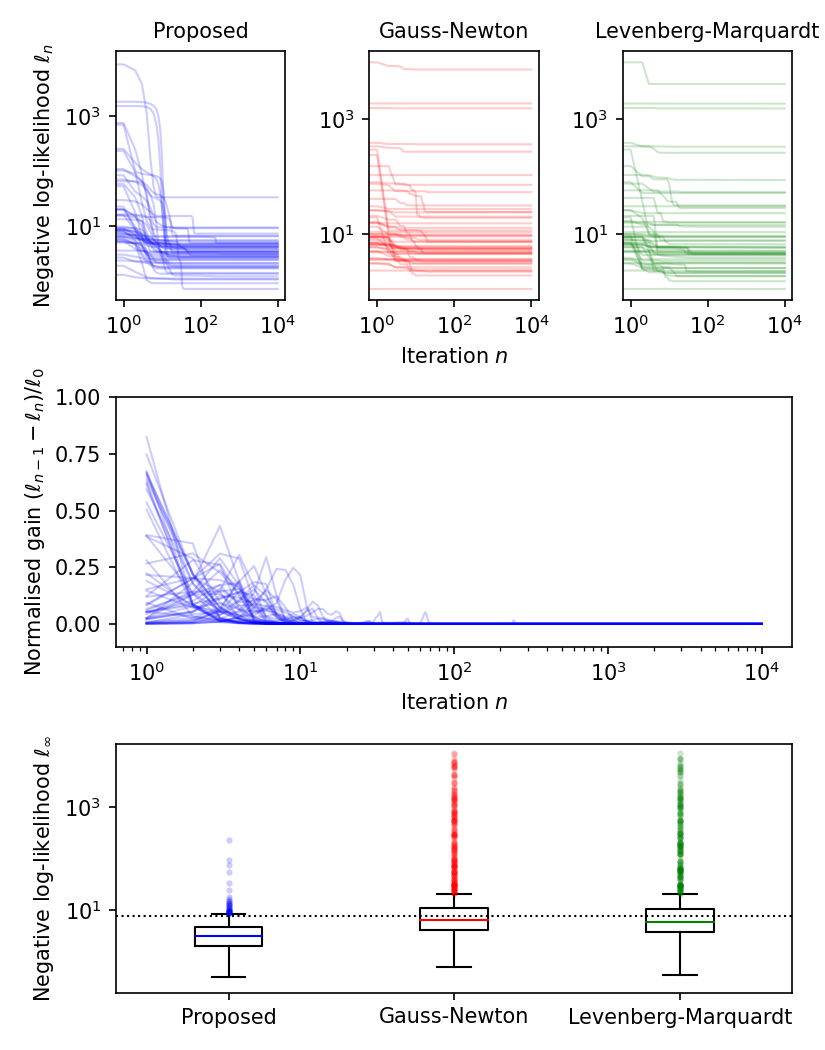}
    \caption{Optimisation of 1,000 simulated voxels for 10,000 iterations using the proposed second order method (left), Gauss-Newton (middle) and Levenberg-Marquardt (right). The top graph shows the negative log-likelihood per iteration for 50 random samples, while the middle graph shows the normalised gain per iteration for these same 50 samples. Gain is only shown for the proposed method. Note that if an iteration was non-monotonic, its gain would be negative, which never happens in our simulations.
    The bottom graph shows the distribution of the final log-likelihood of all 1,000 samples. The dashed line represents the expected log-likelihood of the true parameters. Log-likelihood and iterations are shown in log scale.}
    \label{fig:simu}
\end{figure}

The negative log-likelihood and gain after each fitting iteration in 50 randomly selected voxels are shown in Fig. \ref{fig:simu}. These plots show that all voxels converge monotonically with our approach (the gain is never negative) despite the very wide range of parameters ($\emph{e.g.}$, SNR, observed relaxation times) covered by the simulations. Conversely, Gauss-Newton and Levenberg-Marquardt fail to converge to the optimum in a large number of voxels, despite the fact that this simulation context is very favourable to them: they are allowed to define one regularisation parameter per voxel, whereas in a spatially regularised case, voxels would not be independent and would have to share the same parameter.

Note that under this setup (3 contrasts, 5 echoes per contrasts), the vast majority of these maximum-likelihood fits yield optimal negative log-likelihood that are lower than the expected negative log-likelihood of the true parameters. This shows that ML likely over-fits the data and is strong evidence that regularisation is needed.

\begin{figure}[t!]
    \centering
    \includegraphics[width=\columnwidth]{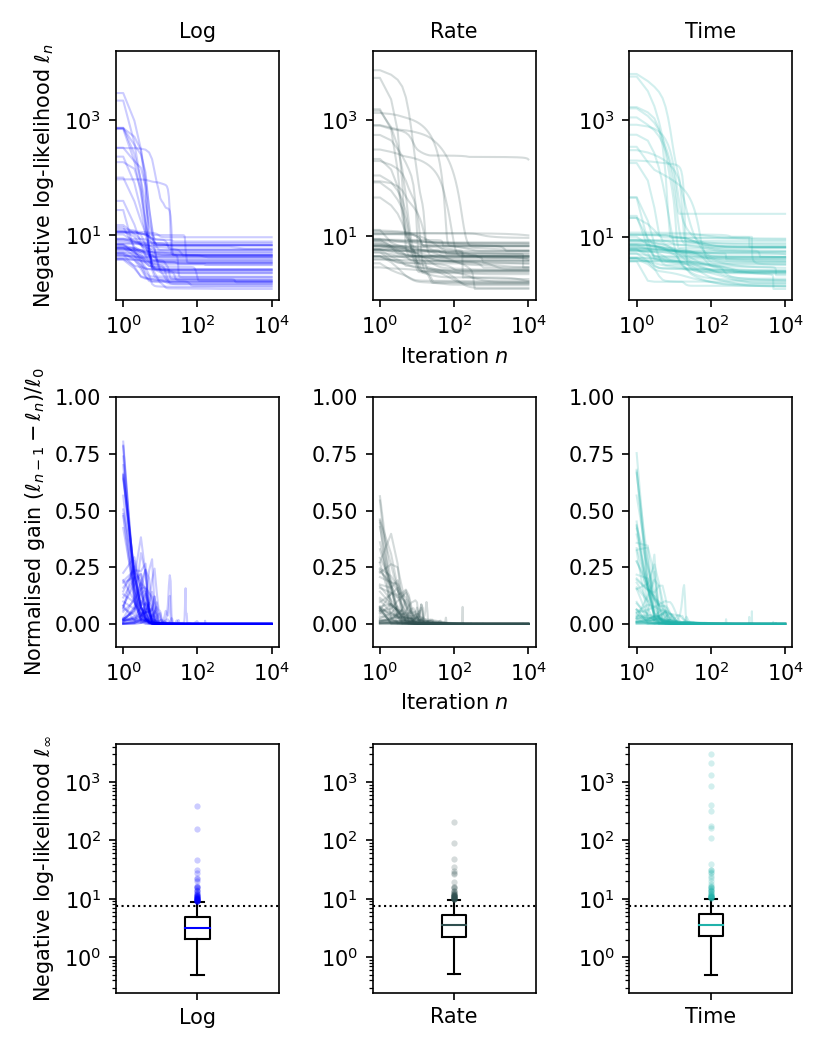}
    \caption{Optimisation of 1,000 simulated voxels for 10,000 iterations using the proposed second order method. The optimisation is either performed in the basis of the log-parameters (left), the rate representation (middle) or the time representation (right). The top graph shows the negative log-likelihood per iteration for 50 random samples, while the middle graph shows the normalised gain per iteration for these same 50 samples.
    The bottom graph shows the distribution of the final log-likelihood of all 1,000 samples. The dashed line represents the expected log-likelihood of the true parameters. Log-likelihood and iterations are shown in log scale.}
    \label{fig:simu_nolog}
\end{figure}

Fig. \ref{fig:simu_nolog} shows results for the same experiment, this time comparing different optimisation basis (log, rate or time). While all methods converge to the same optimum, the rate of convergence is clearly higher when log-parameters are optimised.

\subsection{Hyper-parameter selection}

SPGR+JTV used the same regularisation factor ($\lambda$) for all contrasts. Values $\{1, 5, 10, 15, 20\}$ were tried and the smallest MSE was obtained with $\lambda=10$. Because they have different dynamic scales, NONLIN+JTV used a different factor for the $R_2^\ast$ map ($\lambda_r$) and for the log-intercepts ($\lambda_i$). We tried values $\{0.001, 0.01, 0.1, 1.0, 10\}$ and $\{100, 500, 1000, 5000, 10 000\}$ respectively and found an optimal combination with $\lambda_r=0.1$ and $\lambda_i=1000$. For LOGLIN+AONLM, we used the optimal value $\beta = 0.2$ found in \citet{Balbastre2020}.

\subsection{Methods cross-validation}

\begin{figure}[t!]
    \centering
    \includegraphics[width=\columnwidth]{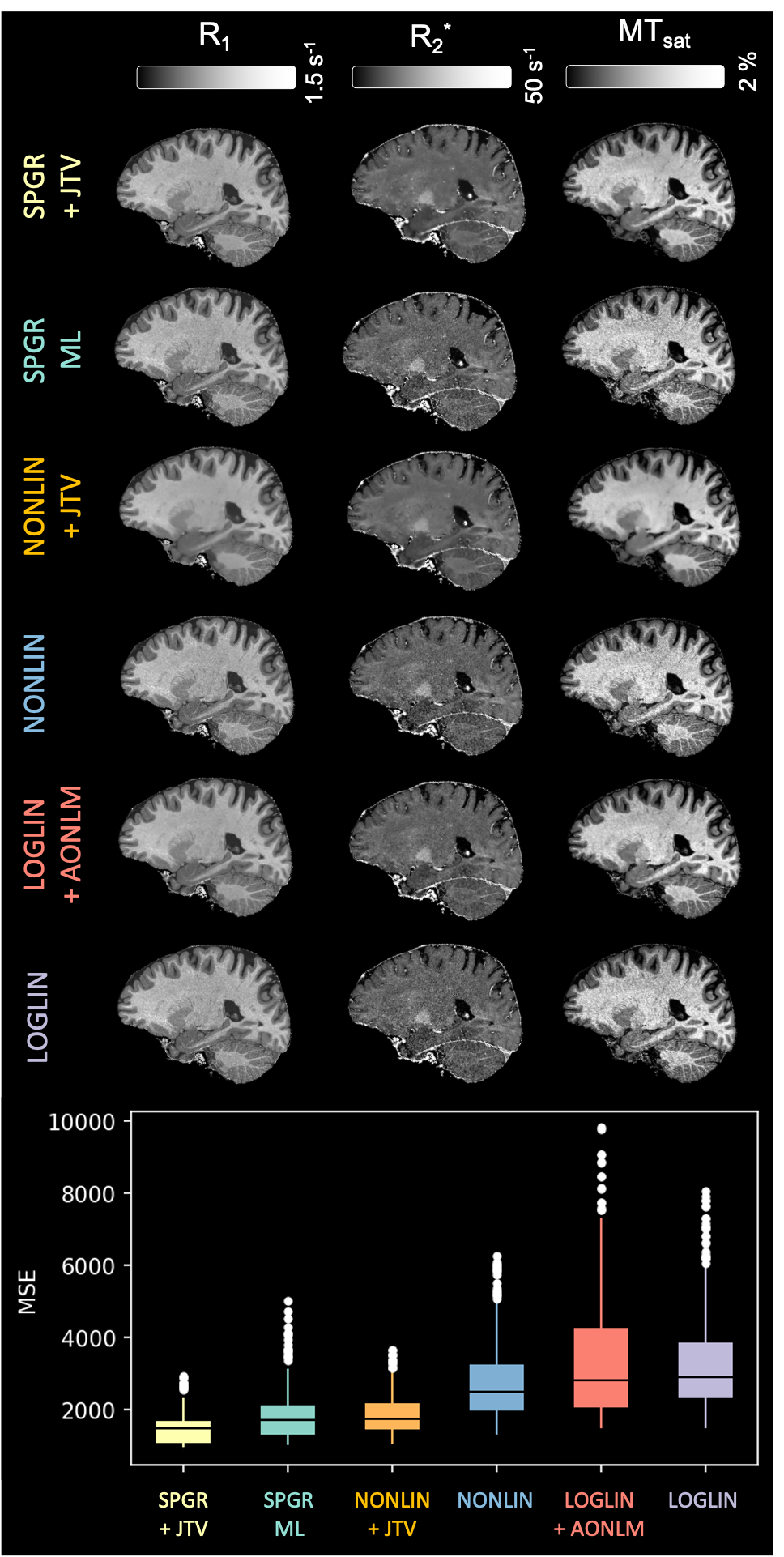}
    \caption{Top: $R_1$, $R_2^\ast$ and MT\textsubscript{sat} maps obtained with all methods. Bottom: z-normalised mean squared error between (unseen) predicted and acquired gradient-echo images. MSE was computed across all voxels in a single volume. Lower is better.}
    \label{fig:maps}
\end{figure}

Quantitative maps obtained with all methods are displayed in Fig. \ref{fig:maps}, along with MSE. As expected, directly fitting the parameters is beneficial over the use of sequential parameter estimation. SPGR+JTV (median MSE = 1459) outperform non-regularised approaches such as SPGR-ML (median MSE = 1706). Methods that make use of rational approximations all trail behind, with the regularised NONLIN+JTV (median 1716) again improving over the non-regularised NONLIN (median 2467). The log-linear version trails far behind (median MSE = 2873), with AONLM preprocessing only slighlty improving it (median MSE = 2799).

\subsection{Echo decimation}

\begin{figure*}[t!]
    \centering
    \includegraphics[width=\textwidth]{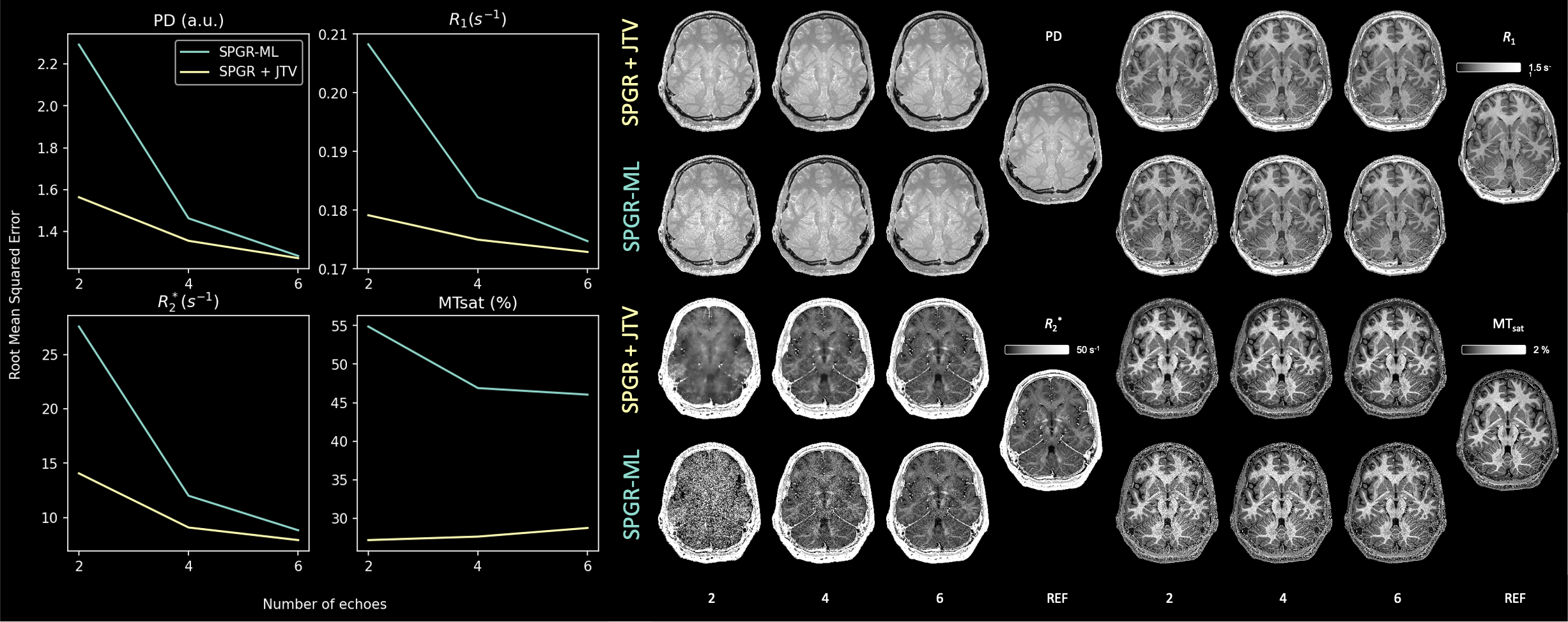}
    \caption{Decimation of the number of echoes. Maps were computed using the first 2, 4 or 6 echoes. Graphs on the left display the mean squared error (lower is better) relative to the reference maps (SPGR-ML with four repeats of 8 echoes) as a function of the number of echoes used. The corresponding parameter maps are displayed on the right.}
    \label{fig:decimation}
\end{figure*}

Fig. \ref{fig:decimation} displays all four parameter maps ($R_1$, $R_2^\ast$, PD, MT\textsubscript{sat}) reconstructed using a growing number of echoes (2, 4, 6) with a regularised algorithm (SPGR+JTV) and a non-regularised one (SPGR-ML). The RMSE between all maps and a reference map (the combined ML solution of all 4 runs) was computed and the regularised maps show a much lower RMSE than the non-regularised ones, even when fewer echoes are used. Noise amplification in the center of the brain appears clearly in the SPGR-ML maps when only two echoes are used, whereas SPGR+JTV manages to preserve anatomical details in this region, despite some additional smoothing. Note that in our implementation, the noise variance and regularisation are two different parameters, even though they share a single degree of freedom in the optimisation. The noise variance is estimated on a volume-by-volume basis while the regularisation factor is kept \emph{a priori} fixed. Consequently, the balance between the likelihood and the prior appears clearly in the maps displayed here, as more smoothing happens when less information (\emph{e.g.}, fewer echoes) is present in the data.

\subsection{Uncertainty mapping}

\begin{figure*}[t!]
    \centering
    \includegraphics[width=\textwidth]{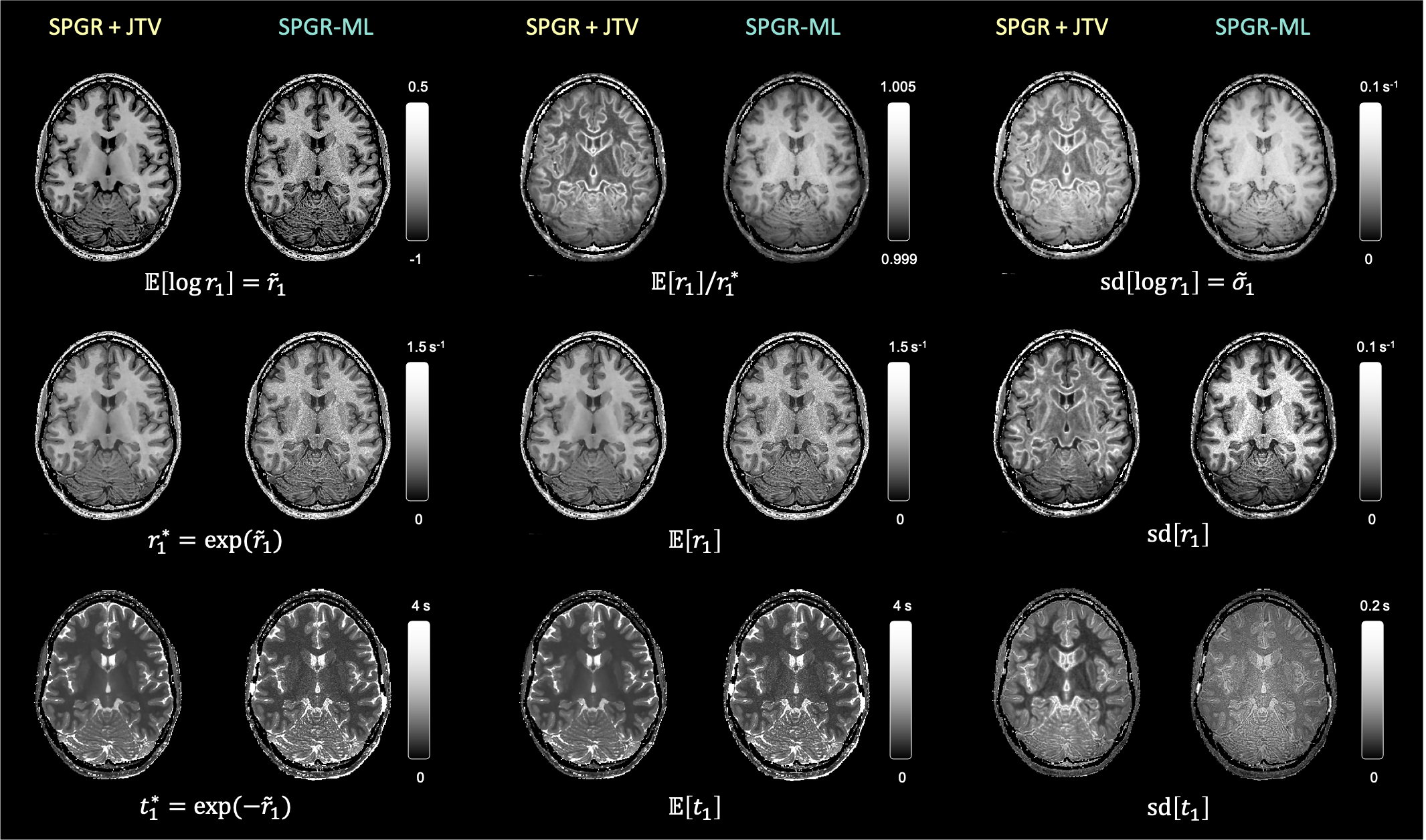}
    \caption{Uncertainty mapping and propagation. Log parameter maps ($\Ex{\log r_1}=\tilde{r}_1$) and their uncertainty ($\operatorname{sd}\left[\log r_1\right]$) were estimated using SPGR+JTV, which includes an edge-preserving prior and SPGR-ML, which has no (or infinitely uninformative) prior. The expected value and standard deviation of $r_1$ and $t_1$ under a Laplace posterior ($\Ex{r_1}$) or a peaked posterior ($\exp(\tilde{r}_1)$) are shown.}
    \label{fig:uncertainty}
\end{figure*}

The capacity to estimate uncertainty is exemplified via $R_1$: posterior expected $R_1$ and $T_1$ maps and their uncertainties are shown in Fig. \ref{fig:uncertainty}. Uncertainty after a SPGR-ML fit is mostly driven by intensities: white matter has a higher uncertainty because its MR signal, and therefore SNR, is generally lower than the grey matter. Conversely, uncertainty after a SPGR+JTV fit is mostly driven by edges, since less local signal averaging is possible in their vicinity.

\section{Discussion \& Conclusion}
\label{sec:conclusion}

In this paper, we introduced an optimisation-based framework for multi-parameter mapping that directly fits the signal equations to all available SPGR data, and can work with (or without) a wide range of regularisers. The key innovation of our method is a second-order optimiser that makes use of a bespoke approximate Hessian that shows excellent experimental convergence. We showed on toy problems that our update steps have theoretical properties that ensure monotonic convergence even when the quadratic approximation is not a majoriser of the objective function. Although we did not offer analytical proof that this monotonic convergence extends to the multivariate case, our solution is at least as good as Gauss-Newton, which is known to converge to the optimum when it converges. Our optimiser therefore  benefits from the same property and, experimentally, was never found to diverge.

We also introduced a novel encoding of the relaxation parameters, based on their log, that makes regularisation insensitive to the choice of unit. This encoding has the additional advantage that parameters live on the entire real line, making the optimisation problem unconstrained.

We evaluated the use of a joint total variation regulariser, an edge-preserving prior that introduces implicit correlations between channels, increasing its denoising power over parameter-wise total variation. We show that this prior yields the best parameter inference, under the metric that inferred parameters best predict unseen data. Although this paper focuses on relatively uninformative spatial regularisers, our method opens the door to the use of informative learned priors. One possibility, inspired by \citet{Brudfors2019d, Brudfors2020}, is to include a multivariate Gaussian mixture model of the log-parameter maps, with learned tissue parameters (means, covariances, tissue probability maps). Alternatively, priors based on learned dictionaries of patches could be used \citep{Dalca2017}.

Although this was not the topic of this paper, it is important to investigate some properties of the maximum-likelihood estimator. While our second-order scheme may output an approximate posterior distribution under the Laplace approximation, it is centered about the mode of the posterior, which is certainly not aligned with the posterior expected value. Different corrections could be devised depending on the map of interest ($\emph{e.g.}$, $R_1$, $T_1$ or $\pm\log R_1$).

The Laplace approximation allowed posterior uncertainty to be estimated. These uncertainty maps are far from perfect: they only contain aleatory uncertainty (\emph{i.e.}, uncertainty that stems from the probabilistic nature of the variables in play, that is known \emph{and} can be modelled), and are based on the assumption that the $B_1^{+/-}$ fields and thermal noise variance are known, thereby discarding any uncertainty about their values. Furthermore, it relies on a Gaussian assumption that could be very wrong. Nonetheless, ‘all models are wrong but some are useful’, and -- while flawed -- these uncertainty maps are better than blind confidence. They could be used to \emph{e.g.} down-weight low-quality data points in downstream regression or classification tasks.

Finally, transmit and receive field maps are assumed pre-computed and fixed in this work and spoiling is assumed perfect. However, it has been shown that key sources of variance in $T_1$ mapping come from imperfect spoiling and errors in the estimate of the $B_1^+$ field \citep{Stikov2015a}. In theory, transmit field map estimation could be integrated in the optimisation framework as long as the corresponding imaging data is included as well \citep{Hurley2012}. However, $B_0$ and $B_1^+$ mapping protocols based on phase differences would be difficult to fit using second-order methods because of the circular support of phase values \citep{Fessler2004}. On the other hand, imperfect spoiling is challenging to integrate in the generative model because, in its current form, it is based on full numerical Bloch simulations and \emph{a posteriori} transformation of \emph{apparent} $T_1$ maps under a polynomial whose coefficients depend on the $B_1^+$ \citep{Preibisch2009b}. Of course, those \emph{post hoc} corrections can still be applied to the parameters estimated here.

Some remaining approximations could also be removed: (1) our Gaussian noise model could be replaced by a more accurate non-central Chi or Rice model that can be fitted using a majorisation-minimisation framework similar to the one used for optimising the JTV functional in this work \citep{Varadarajan2015}; (2) the true TR of the off-resonance pulse could be used instead of being set to zero \citep{Mohammadi2017}, which may remove some residual bias in the $R_1$ and MT\textsubscript{sat} maps. Furthermore, the generative model could be extended to more complex biophysical models, such as multi-compartment models \citep{Deoni2008a}, and to additional sequences, such as spin echo or inversion recovery. Diffusion data could also be integrated and modelled \citep{Jian2007}, as joint diffusion-relaxometry datasets are increasingly being used for accurate microstructure modelling \citep{Hutter2018}. Note that the diffusion signal can also be described using exponential-linear models, which are likely to benefit from our improved Hessian.

On the computational side, this method would benefit from any progress made on TV optimisation. Our implementation uses a relatively simple conjugate-gradient optimiser with Jacobi preconditioning, and it is certain that using preconditioners better tailored to TV would improve its convergence. Additionally, although we experimentally show the stability of our approximate Hessian, we did not provide a proof of convergence in this work. A formal proof that builds on our analysis would surely help move the field forward.

In conclusion, we have presented a flexible probabilistic generative modelling framework, which capitalises on a novel approximate Hessian to enable the robust joint estimation of multiple quantitative MRI parameters. The framework can be adapted to incorporate spatial projections, denoising schemes and additional priors. With this framework we have demonstrated enhanced performance relative to the consecutive estimation techniques commonly used in multi-parameter mapping, with additional benefits such as uncertainty mapping.

\section*{Acknowledgements}

YB, MFC and JA were funded by the MRC and Spinal Research  Charity  through  the  ERA-NET  Neuron  joint  call  (MR/R000050/1). YB was partially supported by the National Institutes of Health under award numbers U01MH117023, R01AG064027 and P41EB030006. CL is supported by an MRC Clinician Scientist award (MR/R006504/1). The  Wellcome  Centre  for  Human  Neuroimaging  is  supported  by  core  funding from the Wellcome [203147/Z/16/Z].


\bibliographystyle{model2-names.bst}\biboptions{authoryear}
\bibliography{main}

\end{document}